\documentclass{article} % For LaTeX2e
\usepackage{iclr2027_conference,times}

\usepackage{amsmath,amsfonts,bm}

\def\eqref#1{equation~\ref{#1}}
\def\1{\bm{1}}

\DeclareMathAlphabet{\mathsfit}{\encodingdefault}{\sfdefault}{m}{sl}
\SetMathAlphabet{\mathsfit}{bold}{\encodingdefault}{\sfdefault}{bx}{n}

\usepackage{hyperref}
\usepackage{url}
\usepackage{graphicx}
\usepackage{booktabs}
\usepackage{multirow}
\usepackage{subcaption}

\title{RAGenome: Scaling Retrieval-Based Genomic Language Models to Long Contexts}

\author{\textbf{Frederikke Isa Marin$^{1}$\thanks{Equal contribution.} \quad
Panagiotis Antoniadis$^{1}$\footnotemark[1] \quad
Dionysia Danai Brilli$^1$} \\
\textbf{Andreas Bjerregaard$^1$ \quad
Rachael DeVries$^{1,2}$ \quad
Yan Li$^1$ \quad
Ole Winther$^{1,3}$\footnotemark[2]} \\
\textbf{Wouter Boomsma$^{1}$}\thanks{Corresponding authors: \texttt{ole.winther@bio.ku.dk, wb@di.ku.dk}} \\[0.5em]
$^1$University of Copenhagen \quad $^2$Novo Nordisk A/S \quad $^3$Technical University of Denmark
}

\iclrfinalcopy % Uncomment for camera-ready version, but NOT for submission.
\begin{document}

\maketitle

\begin{abstract}
The genome holds the blueprint that governs the biological properties of the cell. Consequently, advancing our knowledge of genomic function is crucial both for a broader understanding of biology and for continued biomedical advances. The success of large language models on natural language and protein sequences has motivated similar efforts on genomic data. However, standard genomic language models (gLMs) often require extremely large computational resources and still fall behind traditional methods on some downstream tasks. Recently, MSA-based pretraining has been proposed as an efficient alternative, but existing models are limited to short input contexts, restricting their use to short-range tasks, such as variant effect prediction. In this work, we present RAGenome, the first retrieval-based gLM that scales pretraining to longer contexts (100$\times$ longer than existing MSA-based gLMs), allowing it to capture both across-species evolutionary relationships and within-species longer-range interactions. Trained on whole-genome alignments from 100 vertebrates, RAGenome substantially improves the long-range capabilities of MSA-based gLMs, raising gene finding performance from 0.45 to 0.60, while remaining competitive on purely evolutionary-based tasks like prioritizing pathogenic variants. RAGenome provides competitive gLM performance at a fraction of the training cost, unifying evolutionary modeling and long-range capabilities within a single, flexible, scalable framework. Code is available at \url{https://github.com/PanosAntoniadis/RAGenome}.
\end{abstract}

\section{Introduction}
Advancing our capabilities to decipher and understand genomic sequences, especially the human genome, remains a salient challenge, with broad implications for our understanding of biology and disease \citep{encode2012integrated}. Thanks to advances in sequencing technology, the availability and diversity of genome sequences have increased enormously over the past few decades, but our ability to interpret genomes remains far from complete.

Traditionally, genome modeling has been approached on a per-task basis, like predicting functional annotation, sequence motifs, or variant effects directly from the DNA sequence with task-specific expert models \citep{alipanahi_predicting_2015, zhou2015predicting, kelley2016basset}. In the last decade, protein Language Models (pLMs) have emerged as a highly successful method to extract rich representations from the wealth of unlabeled protein sequences \citep{hayes2025simulating, jain2025e1}, helping to overcome the limitations in availability of protein annotation data. This sparked interest in applying the same language modeling techniques to genomic data, resulting in a variety of genomic Language Models (gLMs) trained on unstructured whole-genome sequences, varying widely in species coverage and context length \citep{ji2021dnabert, nguyen2023hyenadna, dalla2025nucleotide, brixi2026genome}. However, this success comes at a huge training cost: state-of-the-art gLMs like Evo2 require billion-parameter models trained on trillions of nucleotides across many eukaryotic genomes over months of compute \citep{brixi2026genome}. Despite this scale, such large-scale gLMs still fail on certain tasks such as prioritizing pathogenic variants \citep{ye2026gpnstar}.

Another approach is to model evolutionary relationships explicitly by exposing the model to multiple sequence alignments (MSAs) during pretraining, again following prior work on MSA-based protein language models \citep{rao2021msa, jumper2021highly}. While these explicit-homology based models, like GPN-Star \citep{ye2026gpnstar}, have proven very successful on short-context tasks, they cannot scale to long-context tasks. The main reason for this limitation is that they prioritize species depth and process every token of the MSA, including uninformative gap tokens. As a result, their computational cost grows with both context length and the number of species, restricting them to windows of around 128–256 nucleotides. They therefore inherently fail to capture the long-range interactions required for tasks such as gene finding, where splice sites that define exon and intron boundaries can be separated by thousands of base pairs.

Unlike these MSA-based methods, which directly condition on the full alignment, actively selecting the relevant tokens from the available evolutionary information can enable a model to scale efficiently to much longer contexts. We introduce RAGenome (Figure \ref{fig:ragenome}), the first retrieval-based gLM pretrained on homologous sequence context: it retrieves homologous sequences from whole-genome alignments (WGA), scaling to much longer contexts while modeling evolutionary relationships explicitly. To showcase the multiple benefits of our proposed model, we evaluate it on variant effect prediction, which relies primarily on evolutionary modeling, and on gene finding, which tests the long-range capabilities of a model. 
% We view these results as establishing a broader retrieval-based framework that leverages homology-derived signal while extending it to significantly longer contexts.

Our main contributions are:
\begin{itemize}
  \item We develop RAGenome, the first retrieval-based gLM trained on whole-genome alignments from 100 vertebrates to explicitly model evolutionary relationships. RAGenome \textbf{scales to 100$\times$ longer context windows than MSA-based methods} through an efficient retrieval mechanism, using far less compute than large-scale gLMs.
  \item We show that \textbf{extending the input context substantially improves the long-range capabilities} of RAGenome, surpassing GPN-Star \citep{ye2026gpnstar}, the strongest MSA-based gLM, on gene finding (MCC 0.60 vs. 0.45), and narrowing the remaining gap to large-scale, long-context gLMs.
  \item We further demonstrate that RAGenome maintains strong performance on prioritizing pathogenic variants, indicating that \textbf{it efficiently models evolutionary signals despite the long input context}.
\end{itemize}

\section{Background and Preliminaries}

\subsection{Masked Language Modeling for Genomic Sequences}
\label{sec:masked_lm}

A DNA sequence of length $L$ can be represented as a sequence of tokens $x = (x_1, \dots, x_L)$ where  each token corresponds to a nucleotide $x_i \in \{A, C, G, T\}$. To train a gLM using the masked language modeling (MLM) objective \citep{devlin2019bert}, we randomly mask a subset of tokens $M \subset \{1, \dots, L\}$ (usually around 15\% of the total tokens) and then optimize the model to minimize the negative log-likelihood of the masked tokens:
\begin{equation}
    \mathcal{L}_{\text{MLM}} = -\sum_{i \in M} \log p_\theta(x_i \mid x_{\setminus M}),
\end{equation}
where $x_{\setminus M}$ denotes the unmasked tokens. In gLMs, we usually mask short contiguous spans of nucleotides rather than individual positions, encouraging the model to learn dependencies beyond immediately adjacent bases.

\subsection{Multiple Sequence Alignments}
\label{sec:msa}

Multiple sequence alignments (MSAs) have been widely used in training language models especially for protein sequences, offering an explicit evolutionary signal that has proven particularly powerful for modeling structural and functional constraints \citep{jumper2021highly, rao2021msa, jain2025e1}. They are constructed by aligning homologous positions across biological sequences, revealing site-specific patterns of conservation. Extending this idea to entire genomes yields whole-genome alignments (WGAs), which consist of aligned genome assemblies across hundreds of species, enabling evolutionary analysis at genome-wide scale. Formally, a WGA over $N$ species can be represented as a matrix $X \in \mathcal{V}^{N \times L}$, where $L$ is the number of alignment columns and $\mathcal{V} = \{A, C, G, T, \text{-}\}$ is the nucleotide vocabulary extended with a gap token ('-'), which indicates the absence of an aligned nucleotide due to an insertion or deletion in that lineage. In this work, we train RAGenome on the human-referenced WGA of 100 vertebrates from \citet{ye2026gpnstar}, treating the human row as the query and the remaining rows as retrieved homologous sequences.

\section{Related Work}

\subsection{Long-context Genomic Language Models}

Early attempts at training gLMs were direct adaptations of models that succeeded in natural language, such as DNABERT \citep{ji2021dnabert}, in which a BERT model \citep{devlin2019bert} is trained on the human genome using a masked language modeling objective and fixed $k$-mers. However, it soon became evident that increasing the length of the input context is necessary, motivated by the observation that many functional genomic elements, such as gene structures and regulatory interactions, span much longer distances. HyenaDNA \citep{nguyen2023hyenadna} replaces the costly self-attention layers of the transformer with an implicit long-convolution operator and manages to scale context lengths of up to one million base pairs while keeping a single nucleotide resolution. Nucleotide Transformer \citep{dalla2025nucleotide} instead trains an encoder-only transformer on genomes from over 850 species, showing that models trained on multiple species outperform those trained exclusively on the human genome on most downstream tasks, illustrating that modeling evolutionary signals implicitly is beneficial. 
% GENA-LM \citep{fishman2025gena} replaced the full self-attention layer with a sparse BigBird mechanism \citep{zaheer2020big}, reaching input lengths of up to 36,000 base pairs. 
Most recently, Evo2 \citep{brixi2026genome} scaled gLM training across both model parameters, genome coverage and input context. The authors trained a billion-parameter model on 9 trillion nucleotides from a highly curated genomic atlas spanning all domains of life, reaching a 1-million-token context window. It achieved state-of-the-art performance on several downstream tasks, but training required several months on thousands of GPUs.

In a parallel line of work, sequence-to-function models have been proposed, such as AlphaGenome \citep{avsec2026advancing}, which is trained directly on experimental tracks, and NTv3 \citep{boshar2025foundational}, which adds supervised post-training on functional tracks to a gLM. In this work, we focus
on self-supervised pretraining on genomic sequence alone; supervised post-training on functional tracks, as done in NTv3, is orthogonal to our approach and could in principle also be applied to RAGenome.

\subsection{MSA-Based Methods}

Instead of pretraining on genomes from multiple species to implicitly capture evolutionary constraints, another line of work focuses on training gLMs directly on homologous genomic regions across species through MSAs, similar to how AlphaFold \citep{jumper2021highly} leverages protein MSAs to capture structural constraints across residues.

Traditional methods aim to support variant interpretation in humans and other species and include PhastCons \citep{siepel2005evolutionarily}, a phylogenetic hidden Markov model fit on MSAs, and PhyloP \citep{siepel2006new}, a statistical test against a phylogenetic substitution model. GPN-MSA \citep{benegas2025dna} was the first attempt to train a gLM directly on whole-genome alignments (WGAs) of 100 vertebrate species, learning evolutionary signals by masking different parts of the alignment and demonstrating strong performance on non-coding variant effect prediction tasks. Recently, GPN-Star \citep{ye2026gpnstar} extended this framework by using the phylogenetic tree inferred from the MSA to constrain attention between sequences, achieving state-of-the-art performance across a wide range of variant effect prediction tasks in both coding and non-coding regions of the human genome. In a similar direction, Gamba \citep{consens2026predicting} shows that predicting evolutionary rate as an auxiliary pretraining task, rather than retrieving aligned sequences directly, improves performance of small-scale gLMs.

\section{RAGenome}
\label{sec:ragenome}
We propose RAGenome, the first retrieval-based gLM that extends pretraining to long contexts without sacrificing the explicit modeling of evolutionary constraints. We first describe how RAGenome retrieves and processes homologous sequences from a WGA, and then we move on to the architecture and the pretraining process.

\begin{figure}[t]
  \centering
  \includegraphics[width=\textwidth]{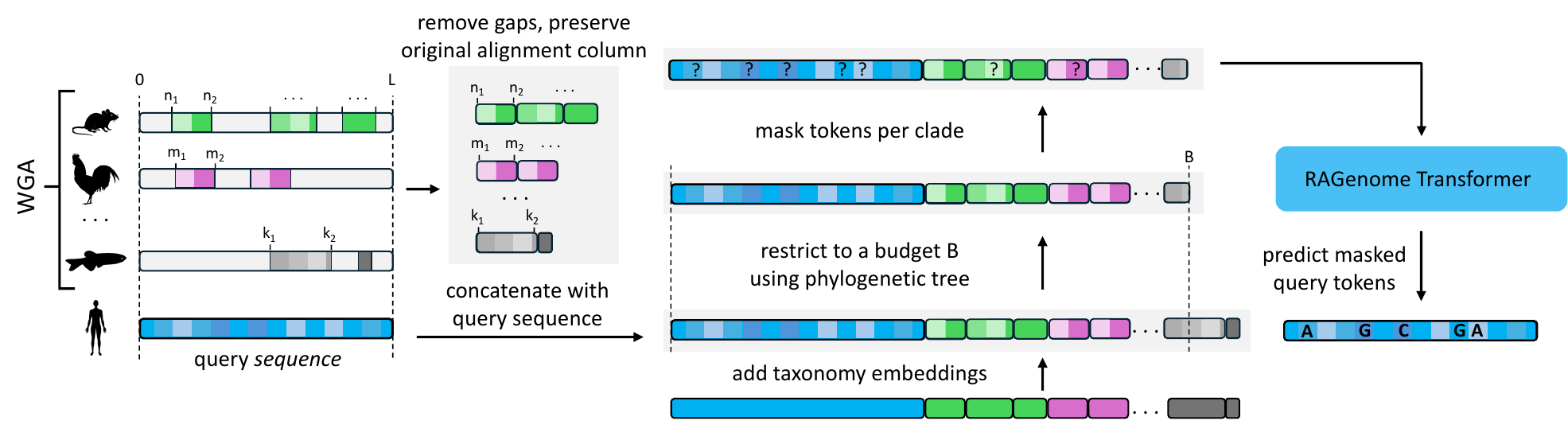}
  \caption{Overview of the retrieval mechanism of RAGenome. For each query sequence, we retrieve aligned homologous sequences from a WGA and remove gap tokens, preserving the original alignment column of each retained token as its positional embedding. We then concatenate all sequences, add taxonomy embeddings to distinguish species, and restrict retrieval to a fixed budget $B$ using the phylogenetic tree. Tokens are then masked per clade, and the resulting sequence is passed to the RAGenome Transformer, which predicts the masked query tokens.}
  \label{fig:ragenome}
\end{figure}

\subsection{Retrieval Mechanism}
\label{sec:retrieval}

As depicted in Figure \ref{fig:ragenome}, RAGenome takes as input a sequence from the query species of length $L$ and the corresponding aligned sequences from a WGA of $N$ vertebrate species. A key inefficiency of current MSA-based gLMs \citep{benegas2025dna, ye2026gpnstar} is that they naively process every input token, while WGAs are highly dominated by uninformative gap tokens that correspond to unaligned segments. Especially for species that are evolutionarily distant from the query, the overall fraction of the human genome that aligns can fall below 2\% \citep{miller200728}. To verify this, we analyzed the WGA of 100 vertebrates used by GPN-Star \citep{ye2026gpnstar} and found that 73\% of tokens correspond to gaps. Note that training a model on only short, highly conserved regions of the genome, as GPN-Star does, is not as inefficient, since conserved regions tend to align well across species and therefore contain fewer gaps (41\% gaps in the top 5\% most conserved regions). However, training a general-purpose foundation model requires pretraining across the full diversity of the genome, where the fraction of gap tokens increases substantially.

\begin{figure}[ht]
  \centering
  \includegraphics[width=\textwidth]{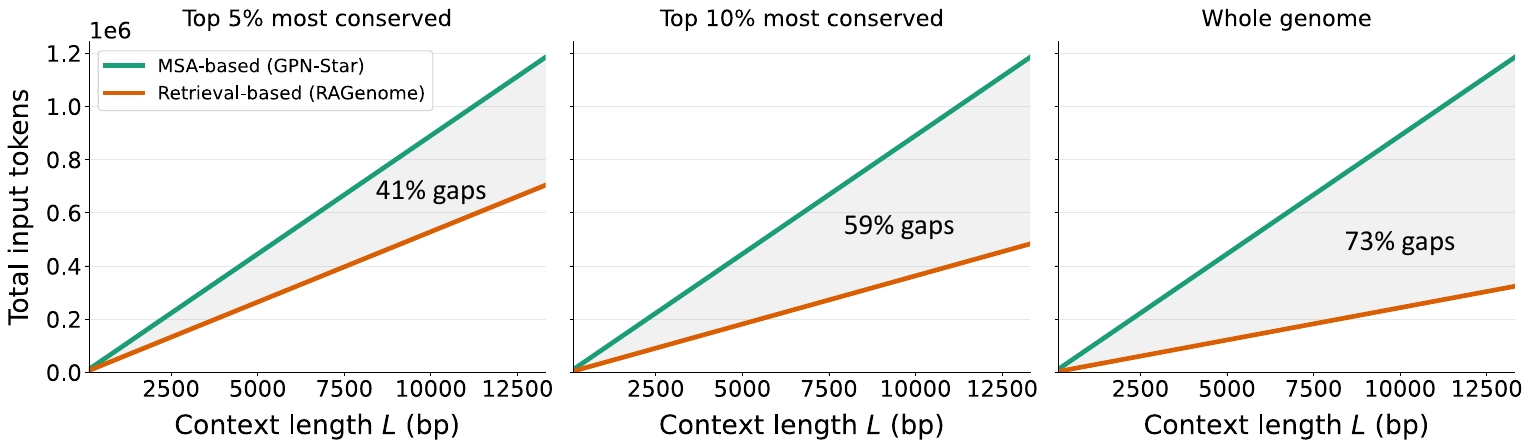}
  \caption{Total number of input tokens as a function of context length $L$, comparing MSA-based methods \citep{ye2026gpnstar} that keep gap tokens against our proposed retrieval-based approach, across three genome coverage levels. The shaded region indicates the reduction in tokens achieved by removing gap tokens. The percentage of gaps in each panel is the total fraction of tokens that are gaps at that coverage level.}
  \label{fig:retrieval_efficiency}
\end{figure}

To make retrieval-based pretraining over the whole genome efficient, we discard gap tokens entirely and keep only aligned nucleotides, which substantially reduces the number of input tokens (Figure \ref{fig:retrieval_efficiency}). \emph{We preserve the original alignment column of each token and use it as its position in rotary position embeddings (RoPE)} \citep{su2024roformer}, applied in both the local and cross-attention layers. In this way, cross-species alignment information is preserved without needing any additional mechanism to track it explicitly. Then, inspired by a similar retrieval technique in pLMs \citep{jain2025e1}, we concatenate all the sequences in a single array and distinguish between them by adding taxonomy embeddings for each species. 

Even after removing the gaps, retaining the full WGA becomes computationally expensive at longer context lengths, since the number of retrieved tokens still grows with both context length and the number of species $N$. We restrict retrieval to a fixed budget $B$, selecting species using the phylogenetic tree to sample sequences from as many distinct lineages as possible within the budget (Appendix \ref{app:species_selection}). We set budget $B$ according to the available GPU capacity since it bounds the memory consumed by the cross-attention layers, whose cost grows with the number of retrieved tokens. For example, in our largest context setting, we set $B=80{,}000$ tokens to fit within an NVIDIA H100 GPU. 

\subsection{Taxonomy Embeddings}
\label{sec:taxonomy}

We include species information in RAGenome by adding taxonomy embeddings to each token representation. Specifically, we map each species $k$ to its NCBI taxonomic lineage, spanning superkingdom, kingdom, phylum, class, order, family, genus, and species \citep{schoch2020ncbi}. Then, we obtain a single shared embedding for each species by concatenating the nominal embeddings of each taxonomy level it belongs to. For example, humans and mice share the same embeddings at the superkingdom, kingdom, phylum and class levels, but differ at the order, family, genus, and species levels, reflecting their more distant evolutionary relationship within mammals. This design enables phylogenetically related species to share common information through their taxonomy embeddings. In Appendix \ref{app:taxonomy_umap}, a UMAP projection \citep{mcinnes2018umap} of the taxonomy embeddings is shown, confirming that phylogenetically related species cluster together in the embedding space.

\subsection{Architecture}
\begin{figure}[t]
  \centering
  \includegraphics[width=\textwidth]{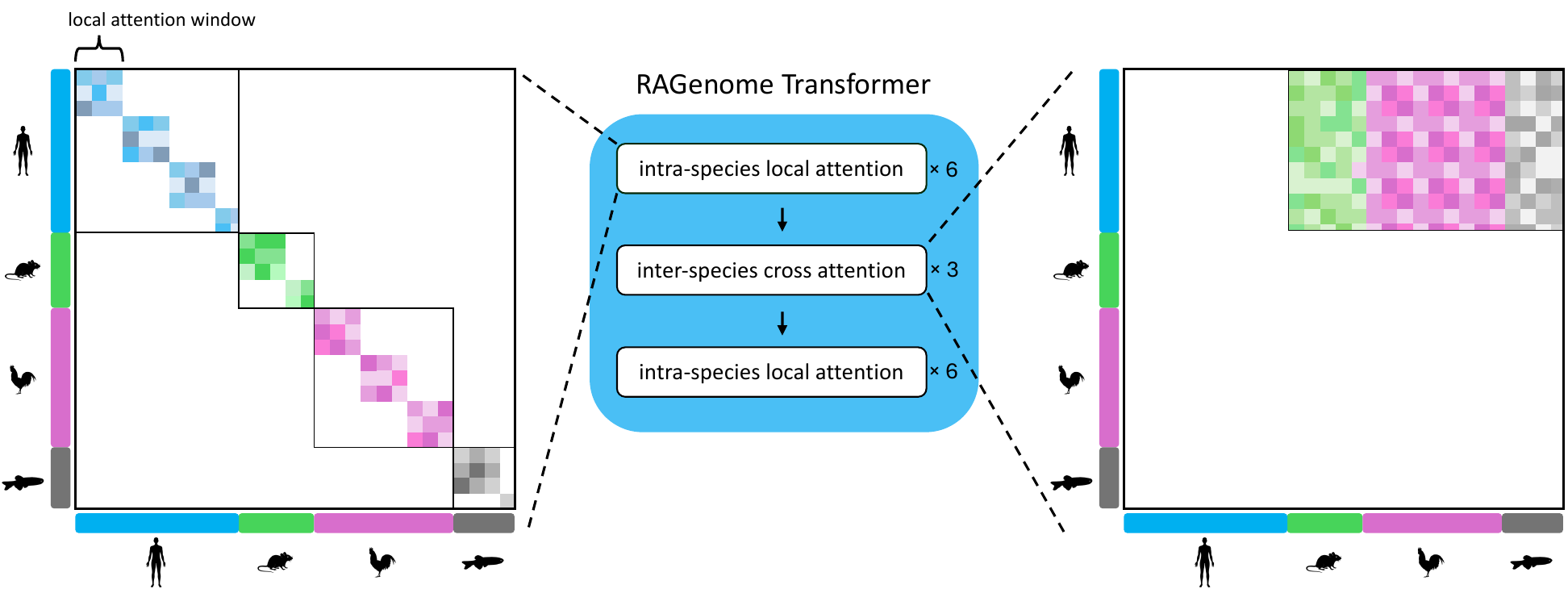}
  \caption{RAGenome Transformer alternates between two types of attention layers. \emph{Left:} Intra-species local attention restricts each position to attend only within a local window neighborhood in the same species. \emph{Right:} Inter-species cross-attention allows the query species (blue) to attend over all the retrieved sequences from the other species. For illustration, local attention windows are shown as non-overlapping blocks.}
  \label{fig:attention}
\end{figure}

RAGenome is a multi-head Transformer \citep{vaswani2017attention} where standard self-attention layers are replaced with two types of attention layers tailored to capture within-species and cross-species dependencies, as shown in Figure~\ref{fig:attention}.

\textbf{Intra-species local attention.} To scale to long sequences while modeling within-species dependencies, we replace full self-attention with local (sliding window) attention \citep{beltagy2020longformer}, where the cost scales linearly instead of quadratically with sequence length. By stacking multiple such layers, the effective receptive field grows with depth, allowing information to propagate across longer parts of the sequence. Let $h^k = (h^k_1, \dots, h^k_{L_k})$ denote the hidden representations of the tokens of species $k \in \{0, 1, \dots, N\}$, where $k=0$ corresponds to the query species. For each species independently, position $i$ attends only to a local window $\mathcal{W}(i) = \{\, j : |i - j| \le w/2 \,\}$ of size $w$:
\begin{equation}
    \text{LocalAttn}(h^k)_i = \text{softmax}\left(\frac{q^k_i (K^k_{\mathcal{W}(i)})^\top}{\sqrt{d}}\right) V^k_{\mathcal{W}(i)},
\end{equation}
where $q^k_i$ is the query vector at position $i$ of species $k$, and $K^k_{\mathcal{W}(i)}$, $V^k_{\mathcal{W}(i)}$ are the keys and values within the window; $i$ indexes the gap-free tokens of species $k$.

\textbf{Inter-species cross-attention.} To enable the model to directly incorporate evolutionary information from the retrieved homologous sequences, we insert inter-species cross-attention layers in the middle of the network. These layers allow information to flow from the retrieved sequences to the query sequence, letting the model learn WGA-dependent representations. Formally, let $h^0_i$ denote the representation of the query species at position $i$ and let $h^{1:N} = \{h^k_j : k \in \{1, \dots, N\}, j \in \{1, \dots, L_k\}\}$ denote the representations from the retrieved $N$ species. Cross-attention at position $i$ is computed as:
\begin{equation}
    \text{CrossAttn}(h^0, h^{1:N})_i = \text{softmax}\left(\frac{q^0_i (K^{1:N})^\top}{\sqrt{d}}\right) V^{1:N},
\end{equation}
where $q^0_i$ is the query vector at position $i$ of the query species, and $K^{1:N}, V^{1:N}$ are the keys and values computed from the concatenated retrieved representations across all species.

\subsection{Pretraining}
\label{sec:pretraining}

During pretraining of RAGenome, we always set \textit{Homo sapiens} as our query species and retrieve homologous sequences from a WGA of 100 vertebrates \citep{ye2026gpnstar}, excluding species in the same clade as the query. This results in 89 species retrieved along with \textit{Homo sapiens}. The model is optimized using a masked language modeling objective where short contiguous spans of nucleotides (3 to 6 nucleotides) are masked together, instead of individual tokens. In each training sample, we randomly select 15\% of the non-gap positions in each query sequence, ensuring that the same positions are selected in species that belong to the same clade.

The target segment length $L$ is set to 13,312 nucleotides, selected as a multiple of the local attention window size $w=512$ that is long enough to capture the gene structures and splice-site interactions. However, directly training a language model on such long sequences can affect training stability as the variance in gradient increases \citep{li2022stability}. Similar to other long-context gLMs \citep{nguyen2023hyenadna, brixi2026genome}, we apply a context extension strategy, where we gradually increase the input context length to improve stability and reduce training time. To account for the retrieval mechanism as well, we also gradually increase genome coverage, from more conserved to noisier, less conserved regions. Specifically, we train RAGenome in four stages (Appendix \ref{app:conserved_regions}):
\begin{enumerate}
    \item \textbf{Short conserved regions.} We start with $L=1,024$ nucleotides restricted to the top 5\% most conserved regions of the human genome, where retrieval signal is most reliable and local motifs are easiest to learn.
    \item \textbf{Coverage extension.} We then extend training to the top 40\% most conserved regions of the genome keeping the same context length, exposing the model to greater sequence diversity and less conserved regions.
    \item \textbf{Context extension.} We extend the input context to $L=4,096$ nucleotides, allowing the model to learn longer-range signals such as gene structure.
    \item \textbf{Full-length context.} Finally, we scale the input context to the full target length of $L=13,312$ nucleotides, covering the entire genome.
\end{enumerate}

We train a 168M-parameter model using AdamW and a learning rate of $10^{-4}$ with a cosine schedule of 4,000 warmup steps at the start of each stage. The full hyperparameters for the architecture and the pretraining stages are shown in Appendix \ref{app:hyperparameters}.

\section{Results}

\subsection{Downstream Tasks}
We evaluate RAGenome on two complementary downstream tasks, testing two capabilities that current gLMs struggle to jointly achieve and that our retrieval-based approach is specifically designed to unify: long-range interaction modeling and evolutionary signal modeling.

\textbf{Gene finding.} Annotating genes and identifying coding sequences requires a model to capture long-range interactions. For example, accurately classifying a given nucleotide as being within an intron or exon depends on information from splice sites that can be located far away. Also, introns vary in length from a few hundred to several thousand base pairs, requiring a language model that can reason over long-range dependencies to make accurate predictions. We evaluate RAGenome on the gene finding task from BEND \citep{marin2024bend}. The task is framed as nucleotide-level 9-class classification: embeddings are extracted from the frozen, pretrained model and used to train a lightweight two-layer CNN classifier on top. For elements longer than the native context of a model, sequences are chunked, embedded independently and concatenated before training the classifier. We report the Matthews Correlation Coefficient (MCC) on the provided test split.

\textbf{Variant effect prediction.} Classifying variants as pathogenic or benign in ClinVar \citep{landrum2014clinvar} is crucial for understanding the impact of genetic variants on disease susceptibility. Classical evolutionary methods such as PhastCons \citep{siepel2005evolutionarily} and PhyloP \citep{siepel2006new} are already known to perform relatively well on this task, indicating that evolutionary conservation is a strong signal for prioritizing pathogenic variants. Following \citet{ye2026gpnstar}, we perform the task zero-shot by computing the log-likelihood ratio between the reference and alternate allele under the model and report the area under the receiver operating characteristic curve (AUROC).

\subsection{Scaling RAGenome Improves Long-Range Capabilities}

\begin{figure}[t]
  \centering
  \includegraphics[width=\textwidth]{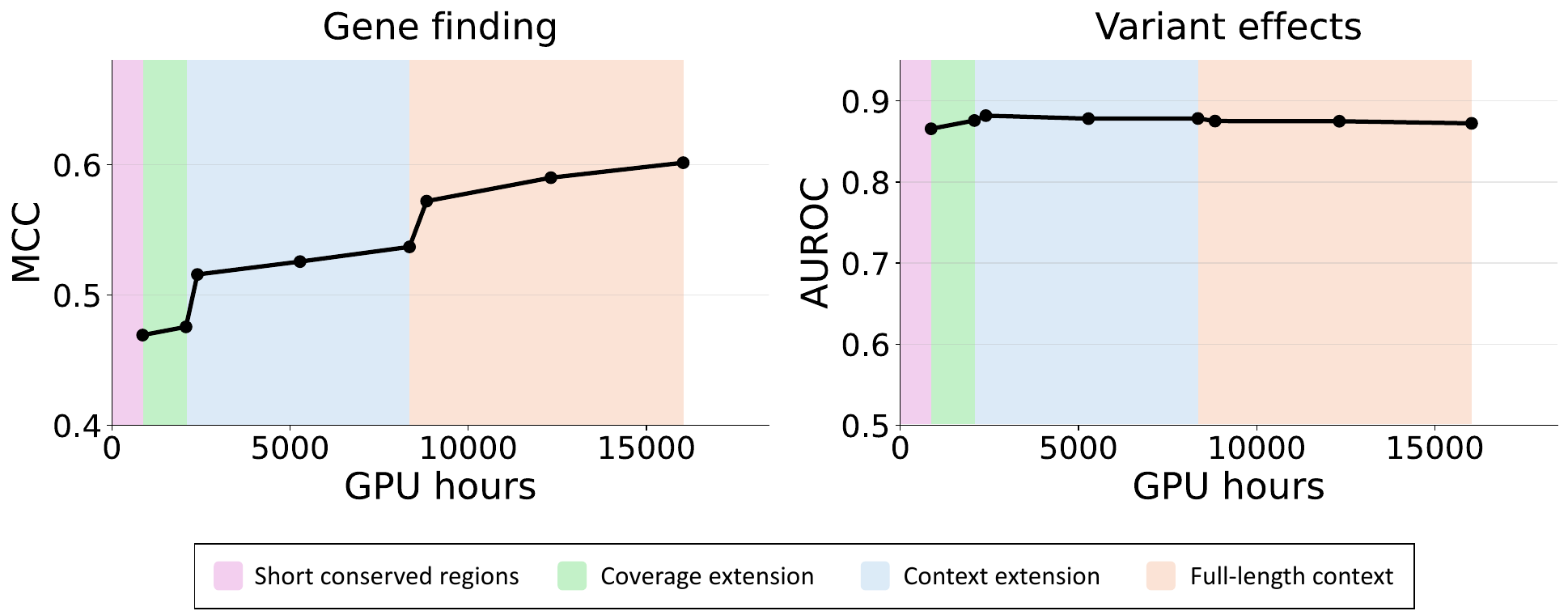}
    \caption{Performance of RAGenome on gene finding (left) and variant effect prediction (right) as a function of training compute. The colored regions indicate the pretraining stages presented in Section \ref{sec:pretraining}. Gene finding performance improves gradually within each stage but increases sharply at each context-extension transition, indicating that longer input context is crucial for gene annotation. Variant effect prediction remains flat, reflecting that it relies primarily on local evolutionary conservation signals captured from the WGA.}
  \label{fig:scaling}
\end{figure}

First, we evaluate the effect of the different pretraining stages discussed in Section \ref{sec:pretraining} on the performance of RAGenome on the two downstream tasks. Figure \ref{fig:scaling} presents gene finding and variant effect prediction performance as a function of pretraining compute, with colored regions indicating each pretraining stage.

We observe that gene finding performance improves substantially after each pretraining stage, with the largest gains occurring immediately after each transition. Specifically, MCC increases from 0.47 to 0.52 after extending the context length to 4,096 nucleotides (blue region), and from 0.54 to 0.57 after further extending it to 13,312 nucleotides (orange region). This suggests that training on longer context windows, rather than training compute alone, is the crucial factor for improving gene finding performance, since the model can capture broader gene structure, such as relationships across splice sites and regulatory elements. In contrast, variant effect prediction remains flat across all pretraining stages, indicating that our retrieval mechanism preserves its strong evolutionary signal even as context length increases. Together, these results indicate that RAGenome does not trade one capability for the other: the same model achieves strong long-range understanding without sacrificing the short-range, evolutionary-driven performance that makes MSA-based gLMs effective. In Appendix \ref{app:context_ablation}, we disentangle the effect of context length from that of additional training compute, 
showing that at a fixed checkpoint, performance improves consistently with inference context length. Similarly, in Appendix \ref{app:budget_ablation} we show an ablation on different retrieval budgets $B$, indicating the importance of retrieval information on both tasks.

\subsection{Comparison with other models}

\begin{table}[t]
  \caption{Results on Gene Finding from BEND \citep{marin2024bend} and Variant Effect Prediction (VEP) from ClinVar \citep{landrum2014clinvar}. Each model is evaluated at its native context length. MSA indicates whether the model uses a whole-genome alignment. For models present in the original BEND paper (HyenaDNA, DNABERT-2, GENA-LM and NT-MS), we evaluated them again ourselves, with only minor differences from the original numbers.}
  \label{tab:comparison}
  \centering
  \begin{tabular}{lcccccc}
    \toprule
    Model            & Parameters & Max length     & MSA        & $\sim$GPU Hours & Gene finding & VEP \\
    \midrule
    GPN-MSA          &  86M          & 128            & \checkmark &  20       &   0.43           & 0.90 \\
    GPN-Star (V)         & 200M       & 128            & \checkmark &  1,000    & 0.45 & 0.92 \\
    \midrule
    HyenaDNA         & 7M      & 1M                &          &  5,376    & 0.33 & 0.51 \\
    DNABERT-2 \textsuperscript{1}   & 117M       & 10,000   &          &  12,000    & 0.41 &  --    \\
    GENA-LM  \textsuperscript{1, 2}  & 336M       & 4,500    &          &  --         & 0.50 &  --   \\
    NT-MS            & 2.5B       & 5,994          &          & 215,000    & 0.66 & 0.57 \\
    Evo2 \textsuperscript{3}  & 7B & 1M             &          & $>$ 1M   & 0.81 & 0.80  \\
    \midrule
    RAGenome  & 168M       & 13,312         & \checkmark &  16,032         & 0.60 & 0.87 \\
    \bottomrule
  \end{tabular}
  \begin{flushleft}
  \small
  \textsuperscript{1} Not applicable for variant effects task due to its tokenization strategy. \\
  \textsuperscript{2} Insufficient information reported to compute GPU-hours. \\
  \textsuperscript{3} GPU-hours estimated from training for several months on over 2,000 GPUs \citep{brixi2026genome}.
  \end{flushleft}
\end{table}

\textbf{Baselines.} RAGenome is an attempt to build a retrieval-based gLM that can model evolutionary signals and long-range interactions at the same time. We compare RAGenome against two families of models: MSA-based models that explicitly leverage homologous sequences and large-scale gLMs that rely on scale and long context instead (Table \ref{tab:comparison}). GPN-MSA \citep{benegas2025dna} and GPN-Star \citep{ye2026gpnstar} are MSA-based models that achieve impressive performance on variant effect prediction, but struggle on gene finding due to their short context length. On the other hand, NT-MS \citep{dalla2025nucleotide} is an attempt to improve performance by scaling model parameters (2.5B) and data coverage (over 850 species), without an MSA. It performs better than GPN-MSA and GPN-Star on gene finding, but performs poorly on variant effect prediction, likely because it does not explicitly model evolutionary conservation. Further scaling in this direction, Evo2 \citep{brixi2026genome}, a 7B-parameter model, achieves strong results on both tasks without an MSA, suggesting that it implicitly learned evolutionary conservation patterns by scaling training across thousands of genomes over several months using over 2,000 GPUs. Finally, the mid-scale long-context models HyenaDNA \citep{nguyen2023hyenadna}, DNABERT-2 \citep{zhou2024dnabert}, and GENA-LM \citep{fishman2025gena} ignore MSAs and underperform on gene finding.

\textbf{Gene finding.} In Table \ref{tab:comparison}, we observe that RAGenome achieves an MCC of 0.60, substantially outperforming both the MSA-based methods (GPN-Star, GPN-MSA) and all the mid-scale gLMs (HyenaDNA, DNABERT-2, and GENA-LM). This supports that identifying genomic elements benefits from long-range context, since gene structure depends on relationships across splice sites and distant regulatory elements that a short context window cannot capture. However, context length alone is not sufficient: HyenaDNA supports contexts of up to 1M nucleotides, but still performs poorly (MCC 0.33), indicating that sufficient training compute is also essential for learning long-range complex relationships. RAGenome underperforms relative to the two much larger-scale gLMs, NT-MS (0.66) and Evo2 (0.81), which are trained with far more parameters and far more compute, likely explaining the remaining gap. Given that the performance of RAGenome on gene finding also continues to improve with longer context (Figure \ref{fig:scaling}), we expect further scaling in model size, training data, and context length to close this gap.

\begin{figure}[t]
  \centering
  \includegraphics[width=\textwidth]{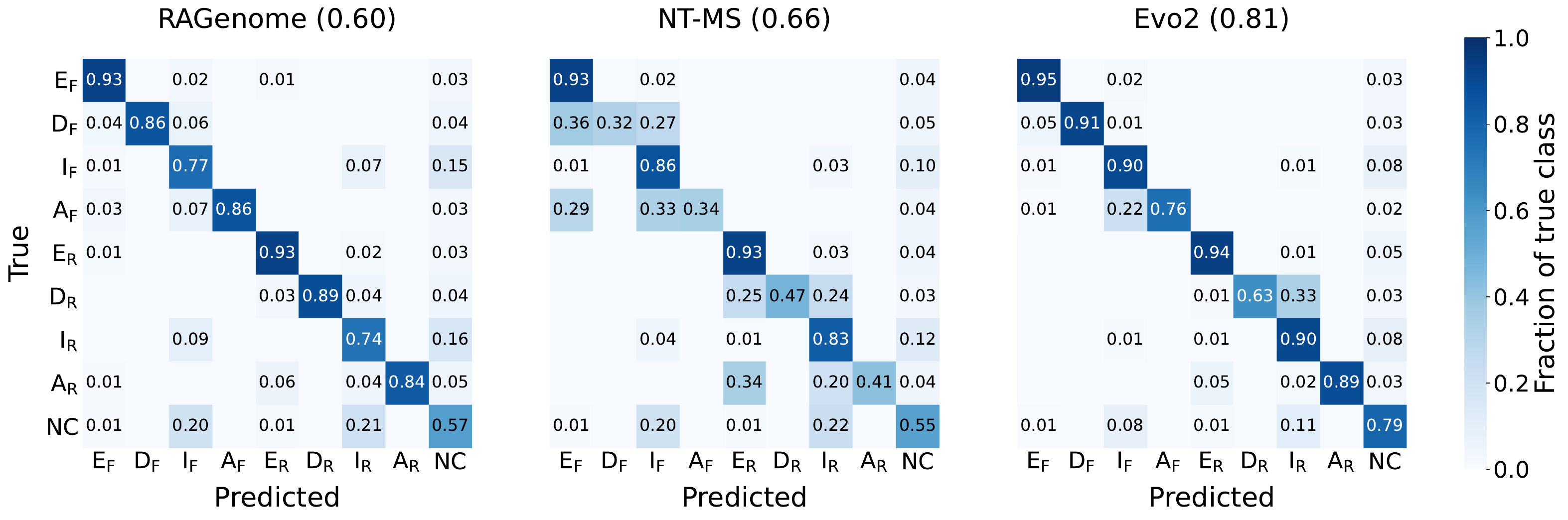}
    \caption{Per-class confusion matrices for RAGenome, NT-MS, and Evo2 on the gene finding task \citep{marin2024bend}, normalized by true class. RAGenome outperforms NT-MS at identifying donor (D) and acceptor (A) splice sites on both the forward (F) and reverse (R) strand, while NT-MS is better at classifying introns (I). Evo2 outperforms both models on most classes given its much larger scale in model size, data, and compute.}
  \label{fig:confusion_matrix}
\end{figure}

To further analyze the performance comparison with large-scale gLMs, Figure \ref{fig:confusion_matrix} shows per-class confusion matrices for RAGenome, NT-MS, and Evo2 on the gene finding task. We observe that RAGenome substantially outperforms NT-MS at identifying donor and acceptor splice sites, while NT-MS is better only at introns. Although RAGenome matches or outperforms NT-MS on most classes, the aggregate MCC of NT-MS is higher because introns span far more nucleotide positions in the test set. Evo2 outperforms both models on most classes, reflecting its much larger scale in model size, data, and compute. Together, these results indicate that our retrieval-based approach can narrow the gap to large-scale gLMs using substantially fewer parameters and less compute.

\textbf{Variant effect prediction.} On variant effect prediction, RAGenome achieves an AUROC of 0.87, surpassed only by the two MSA-based models GPN-MSA (0.90) and GPN-Star (0.92), that are specifically optimized for this short-range, evolutionary-driven task, while it substantially outperforms both large-scale gLMs, Evo2 (0.80) and NT-MS (0.57).

Altogether, if we set aside Evo2, whose model size and training compute set it in a different regime from every other baseline, RAGenome is the only model in Table \ref{tab:comparison} that performs competitively on both tasks, indicating that a retrieval-based gLM can unify evolutionary modeling and long-range capabilities within a single, efficient architecture.

\section{Limitations}

RAGenome requires a WGA at inference and cannot currently be applied to sequences or species for which no alignment of any sort is available. As future work, we plan to address this by training a single model that works both with and without an alignment. Additionally, RAGenome is trained with \emph{Homo sapiens} as the query species, and its behavior on other species remains untested for now. While the architecture supports any query species and alignment, applying it to a new one would likely require additional pretraining. Finally, we trained a single 168M-parameter RAGenome model and leave scaling with model size and data to future work.

\section{Conclusion}
In this work, we investigated how retrieval-based methods, previously proposed for pLMs \citep{jain2025e1}, can be applied to gLMs in order to scale evolutionary modeling to long contexts. We presented RAGenome, the first retrieval-based gLM that scales to input context up to 13,312 nucleotides (100$\times$ longer than MSA-based methods) by using an efficient retrieval mechanism that removes gap tokens from the MSA. RAGenome surpasses GPN-Star, the strongest MSA-based gLM, on gene finding, demonstrating strong long-range capabilities. At the same time, it outperforms much larger gLMs such as NT-MS and Evo2 on variant effect prediction by explicitly modeling evolutionary signals through an MSA. Together, these results suggest that RAGenome is the first gLM to unify explicit evolutionary modeling with long-range genomic understanding within a single, efficient architecture. We expect that further scaling of RAGenome in model size, training data and context length will continue to close the remaining gap to large-scale gLMs, while preserving the efficiency benefits of a much smaller, retrieval-based approach.

\subsection*{AI use statement}
In this work, we used generative AI tools for drafting and editing text throughout the manuscript and suggesting paper structure, titles, and keywords. We have not used generative AI tools to develop theoretical models or conceptual frameworks, propose or refine hypotheses, design or give feedback on research methodology or experiments, implement methods, or support qualitative or thematic data analysis. We have reviewed all AI-assisted work: all experimental results, tables, and figures were produced independently by the authors, and all AI-suggested text was manually verified against primary sources before inclusion. We take responsibility for the final content of this work, including text, claims, and artifacts produced with the aid of generative AI.

\subsection*{Ethics statement}
This work uses only publicly available genomic data: the human reference genome, a whole-genome alignment \citep{ye2026gpnstar}, and the public BEND \citep{marin2024bend} and ClinVar \citep{landrum2014clinvar} benchmarks, none of which contain individually identifiable data.

\subsection*{Reproducibility statement}
In Section \ref{sec:ragenome}, we present in detail the retrieval mechanism, architecture, and pretraining procedure of RAGenome, with full hyperparameters in Appendix \ref{app:hyperparameters}. All datasets for training and evaluating the model are publicly available. Code is available at \url{https://github.com/PanosAntoniadis/RAGenome} and the trained model weights at \url{https://huggingface.co/pantoniadis/RAGenome}.

\subsubsection*{Acknowledgments}
Model training and inference was made possible on Gefion supercomputer by a pilot project from DCAI and a voucher grant awarded by the Novo Nordisk Foundation (NNF25OC0105165). Further support was provided from the Novo Nordisk Foundation through the MLSS Center (Basic Machine Learning Research in Life Science, NNF20OC0062606), CAZAI (DDB and OW, NNF22OC0077058), NovoSTAR programme (RD), and the Pioneer Centre for AI (DNRF grant number P1). We also thank Anders Krogh for invaluable discussions.

\bibliography{references}
\bibliographystyle{iclr2027_conference}

\newpage
\appendix
\section{Appendix}

\subsection{Species Selection under the Budget}
\label{app:species_selection}

As described in Section \ref{sec:retrieval}, we restrict retrieved tokens to a fixed budget $B$ by selecting species using the phylogenetic tree. We use the same phylogenetic clades as GPN-Star \citep{ye2026gpnstar}, excluding the clade of the query species. For each training sample, we shuffle the clades and draw one random species from each clade without replacement; we then repeat this process, drawing an additional species from each clade that still has one remaining, until the number of aligned tokens reaches the budget $B$. This promotes diversity across clades, since every clade contributes one species before any clade contributes a second.

\subsection{Taxonomy Embeddings}
\label{app:taxonomy_umap}

\begin{figure}[ht]
\centering
\begin{subfigure}[b]{0.48\textwidth}
    \centering
    \includegraphics[width=\textwidth]{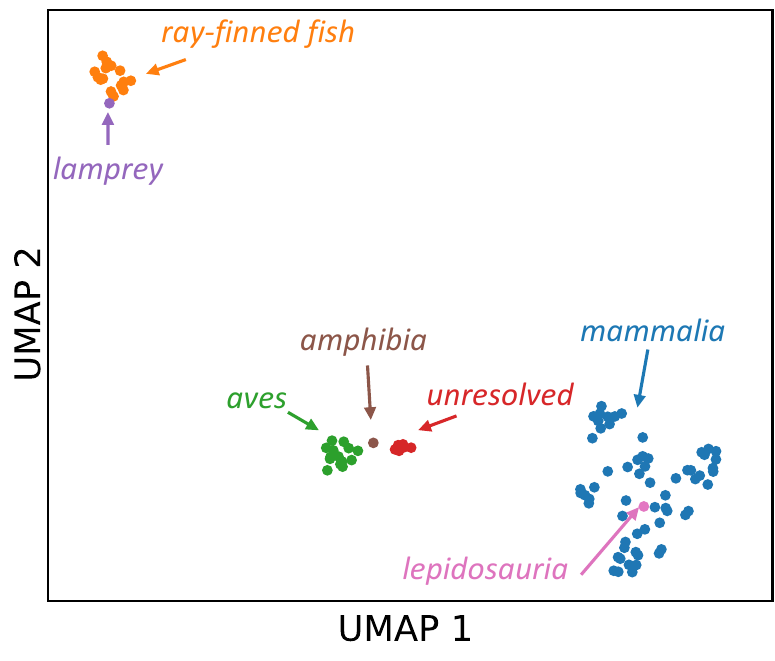}
    \caption{All 100 vertebrates colored by class rank in the taxonomy.}
    \label{fig:taxonomy_umap_class}
\end{subfigure}
\hfill
\begin{subfigure}[b]{0.48\textwidth}
    \centering
    \includegraphics[width=\textwidth]{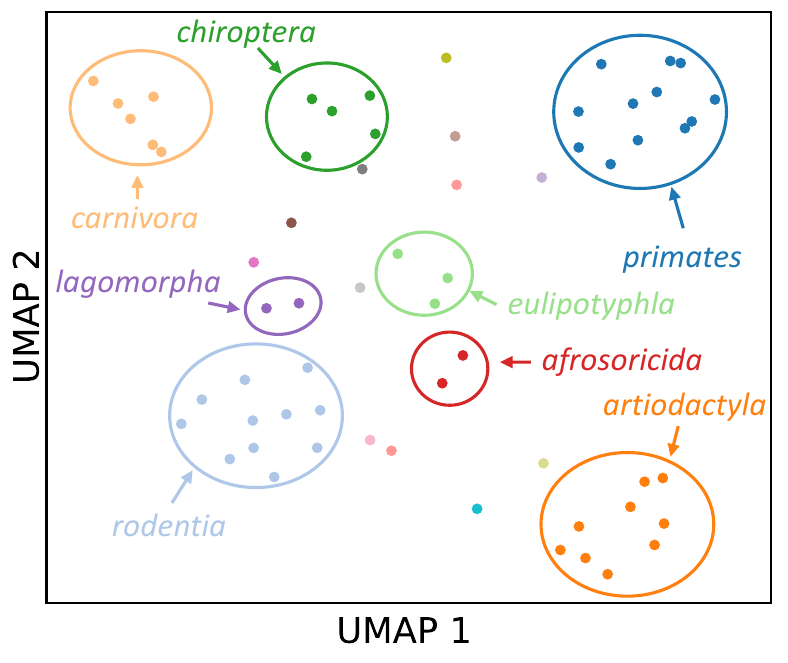}
    \caption{Species of Mammalia class colored by order rank in the taxonomy.}
    \label{fig:taxonomy_umap_order}
\end{subfigure}
\caption{UMAP projection of the learned taxonomy embeddings for the 100 vertebrate species of the WGA.}
\label{fig:taxonomy_umap}
\end{figure}

As described in Section \ref{sec:taxonomy}, we include taxonomy embeddings in RAGenome by concatenating nominal embeddings of its NCBI taxonomic lineage. To check whether the taxonomy embeddings contain phylogenetically related information, we project the learned embeddings of the 100 vertebrate species in our WGA using UMAP \citep{mcinnes2018umap} and illustrate the first two dimensions in Figure \ref{fig:taxonomy_umap}. We clearly observe that species belonging to the same class rank form well-separated clusters (Figure \ref{fig:taxonomy_umap_class}). Within the largest class, Mammalia, species further separate by order rank (Figure~\ref{fig:taxonomy_umap_order}) confirming that taxonomic information is encoded in the learned embeddings.

\subsection{Selection of Conserved Regions}
\label{app:conserved_regions}

When pretraining RAGenome, we first warm up the model on short segments from
highly conserved regions of the genome and then gradually move to longer and
less conserved regions. Following GPN-Star \citep{ye2026gpnstar}, we split the human genome into fixed-size windows and rank them by conservation using PhastCons scores \citep{siepel2005evolutionarily}. Using a window of size $L$, we slide across the genome with a stride of $\frac{L}2$ and assign each window a conservation score equal to the 75th percentile of the PhastCons scores of its bases, discarding windows in which more than 50\% of the bases have no PhastCons score. We then keep a fraction $q$ of windows with the highest scores, plus a random sample of 0.1\% of the remaining windows, so that less conserved sequence is also seen. We use $q=5\%$ for stage 1 and $q=40\%$ for stages 2 and 3, and no selection for stage 4, where all valid windows are used.

\newpage
\subsection{Hyperparameters}
\label{app:hyperparameters}

\begin{table}[ht]
\centering
\caption{Hyperparameters of RAGenome.}
\label{tab:ragenome_hyperparameters}
\small
\begin{tabular}{ll}
\toprule
\multicolumn{2}{l}{\textbf{Architecture}} \\
\midrule
Parameters & 168M \\
Hidden dimension & 1024 \\
Attention heads  & 16 \\
Dimension per head & 64 \\
Transformer Layers & 15 (12 local attention, 3 cross-attention) \\
Local attention ($w$) & 512 \\
RoPE embeddings dimension & 32\\
Activation function  & GELU  \\
\midrule
\multicolumn{2}{l}{\textbf{Optimization}} \\
\midrule
Optimizer & AdamW  \\
Learning rate & $10^{-4}$ \\
Learning rate Schedule & cosine, 4,000 warmup steps \\
Weight decay &  0.01 \\
Gradient clipping & 1.0 \\
Precision & bf16 mixed \\
\bottomrule
\end{tabular}
\end{table}

\begin{table}[ht]
\centering
\caption{Hyperparameters of pretraining strategy.}
\label{tab:stage_hyperparameters}
\small
\begin{tabular}{lcccc}
\toprule
 & Stage 1 & Stage 2 & Stage 3 & Stage 4  \\
\midrule
Context length $L$ & 1,024 & 1,024 & 4,096 & 13,312  \\
Genome coverage & top 5\% & top 40\% & top 40\% & full  \\
Retrieval budget $B$ & 24,000 & 24,000 & 80,000 & 80,000  \\
Effective batch size & 256 & 512 & 512 & 512  \\
GPU hours & 864 &  1,216 & 6,272 & 7,680  \\
\bottomrule
\end{tabular}
\end{table}

\subsection{Ablation on inference context length}
\label{app:context_ablation}

To isolate the effect of context length from the confound of additional pretraining compute, we evaluate RAGenome at different context lengths during inference (Table \ref{tab:context_ablation}). For each pretraining stage, we take the final checkpoint and evaluate it both at its training context length and at the shorter context lengths used in earlier training stages. We observe that across all checkpoints, performance improves monotonically with inference context length, and the fully-scaled model, trained and evaluated at 13,312 nucleotides, achieves the best performance. These results indicate that both additional training and longer context contribute, and that the model effectively uses long-range information at inference.

\begin{table}[h]
  \centering
  \begin{tabular}{lcc}
    \toprule
    Train context & Inference context &  Gene finding (MCC) \\
    \midrule
    1,024  & 1,024   & 0.47 \\
    \midrule
    \multirow{2}{*}{4,096} & 1,024   & 0.51 \\
                            & 4,096  & 0.54 \\
    \midrule
    \multirow{3}{*}{13,312} & 1,024    & 0.53 \\
                             & 4,096    & 0.57 \\
                             & 13,312  & 0.60 \\
    \bottomrule
  \end{tabular}
  \caption{Gene finding performance of RAGenome across different pretraining stages, evaluated at different context lengths.}
  \label{tab:context_ablation}
\end{table}

\subsection{Ablation on Retrieval Budget}
\label{app:budget_ablation}

\begin{figure}[ht]
  \centering
  \includegraphics[width=\textwidth]{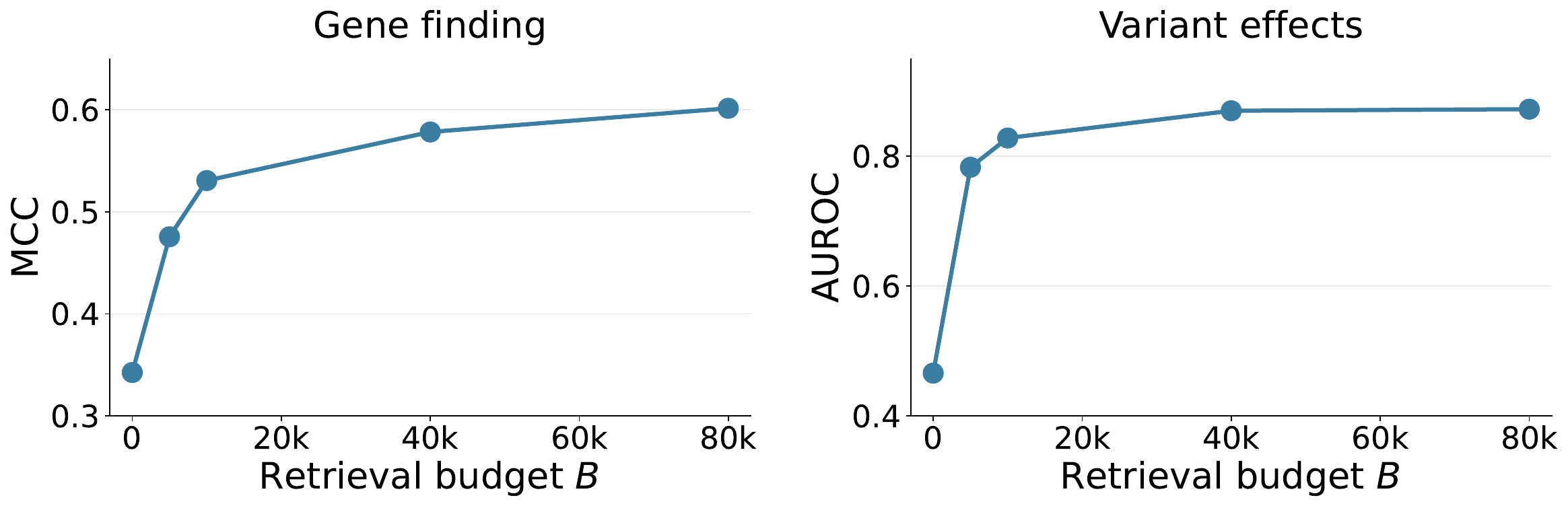}
  \caption{Performance of RAGenome on gene finding (MCC, left) and variant effect prediction (AUROC, right), evaluated at different retrieval budgets $B$ at inference time.}
  \label{fig:budget_ablation}
\end{figure}

To evaluate the effect of the retrieval budget $B$, we evaluate RAGenome with different budgets at inference, from 0 (no retrieved sequences) up to its training budget (Figure \ref{fig:budget_ablation}). Without retrieval, performance collapses, which is expected since the model was never trained without retrieved context. Performance improves substantially as soon as a small budget is used. Variant effect prediction then saturates by $B=40{,}000$, while gene finding improves slightly up to the full budget. This suggests that variant effect prediction relies on local conservation signals that a moderate number of retrieved species already provides, while gene finding continues to benefit from additional evolutionary context.

\end{document}